\documentclass[10pt, a4paper]{article}
\usepackage{lrec2022} % this is the new LREC2022 Style
\usepackage{multibib}
\newcites{languageresource}{Language Resources}
\usepackage{graphicx}
\usepackage{tabularx}
\usepackage{soul}
\usepackage{titlesec}
\titleformat{\section}{\normalfont\large\bfseries\center}{\thesection.}{1em}{}
\titleformat{\subsection}{\normalfont\SmallTitleFont\bfseries\raggedright}{\thesubsection.}{1em}{}
\titleformat{\subsubsection}{\normalfont\normalsize\bfseries\raggedright}{\thesubsubsection.}{1em}{}
\renewcommand\thesection{\arabic{section}}
\renewcommand\thesubsection{\thesection.\arabic{subsection}}
\renewcommand\thesubsubsection{\thesubsection.\arabic{subsubsection}}
\usepackage{epstopdf}
\usepackage[utf8]{inputenc}

\usepackage{hyperref}
\usepackage{xstring}

\usepackage{color}

\title{APPReddit: a Corpus of Reddit Posts Annotated for Appraisal}

\name{Marco Antonio Stranisci$^{\ast}$, Simona Frenda$^{\ast\diamond}$, Eleonora Ceccaldi$^{\dagger}$, \\ \bf \large Valerio Basile$^{\ast}$, Rossana Damiano$^{\ast}$, Viviana Patti$^{\ast}$} 

\address{$^{\ast}$University of Turin - Departiment of Computer Science \\ C.so Svizzera 185, 10149, Turin, Italy. \\
$^{\diamond}$Universitat Polit\`ecnica de Val\`encia - PRHLT Research Center, Valencia, Spain\\
\{marcoantonio.stranisci,simona.frenda,valerio.basile,rossana.damiano,viviana.patti\}@unito.it
\\
$^{\dagger}$Casa Paganini - InfoMus, DIBRIS, University of Genoa \\ P.zza S. Maria in Passione 34, 16123, Genoa, Italy. \\ eleonoraceccaldi@gmail.com}

\abstract{Despite the large number of computational resources for emotion recognition, there is a lack of data sets relying on appraisal models. According to Appraisal theories, emotions are the outcome of a multi-dimensional evaluation of events. In this paper, we present APPReddit, the first corpus of non-experimental data annotated according to this theory. After describing its development, we compare our resource with enISEAR, a corpus of events created in an experimental setting and annotated for appraisal. Results show that the two corpora can be mapped notwithstanding different typologies of data and annotations schemes. A SVM model trained on APPReddit predicts four appraisal dimensions without significant loss. Merging both corpora in a single training set increases the prediction of $3$ out of $4$ dimensions. Such findings pave the way to a better performing classification model for appraisal prediction.
 \\ \newline \Keywords{Emotion Recognition, Appraisal Theories, Annotated Corpora}}

\begin{document}

\maketitleabstract

\section{Introduction}
%Affective computing \cite{picard2000affective} is an %mportant research issue in NLP. 
In the last 15 years, several corpora and lexicons have been developed  with the aim of deepening the analysis of sentiment in texts~\citelanguageresource{strapparava2007semeval,warriner2013norms,cambria2010senticnet}. These resources are heterogeneous, reflecting the high variety of emotion theories in psychological literature (see:~\newcite{sander2018appraisal} for an extensive review). A first group \citelanguageresource{cambria2010senticnet,bertola2016ontology,li2017dailydialog} relies on categorical emotion models, according to which there 
%are 
is a set of basic emotions emerged within the evolutionary process~\cite{plutchik1991emotions,ekman1992there}. Another group of computational resources \citelanguageresource{warriner2013norms,buechel2017emobank,mohammad2018obtaining} refers to dimensional models, which focus on three independent cognitive dimensions along which emotions are evaluated and mapped: Valence, Arousal, and Dominance \cite{russell1980circumplex,bradley2007emotion}. 

NLP resources modeled on appraisal theories, on the other hand, are still missing, with the exception of~\newcite{hofmann2020appraisal} and~\newcite{scherer1994evidence}.
Theories of appraisal~\cite{smith1985patterns,roseman2001appraisal} emphasize the evaluation stage of an event or a situation, that leads to an emotional response and to a corresponding behavior aimed at coping with the situation and alleviating the response itself.
Emotions, in this view, stem from cognitive evaluations of events and are followed by specific autonomic responses, behavioral configurations and action tendencies~\cite{smith1985patterns}. Such evaluations work by assessing the current situation against a set of appraisal criteria, such as the congruence between an event and an agent's goal or the novelty of a specific situation \cite{sander2018appraisal}. Different evaluations of the same situation elicit different emotions. For instance, a given  event will cause joy if it helps the appraising agent fulfill their goal and will elicit surprise if unexpected. 
%These theories define and detail the evaluation processes and response configurations for each emotion \cite{sander2018appraisal}. 
While appraisal theories have been largely employed in computational models of behavior \cite{marsella2009ema,dias2014fatima}, there is still a lack of linguistic resources drawing from this family of theories.  
Appraisal-based linguistic resources may be of great importance because they define, beyond emotions, evaluation processes for situation types and, even more importantly, a range of corresponding  behaviors, of which linguistic behaviors are a subset. Together, all these features could provide more information and explanatory capacity to several tasks like stance detection, abusive language identification, and sentiment analysis. 
%Unfortunately, there is a lack of linguistic resources based on this model. 
Recently, the enISEAR corpus \citelanguageresource{hofmann2020appraisal} has paved the way to the creation of resources which account for emotional appraisal. However, since the corpus developed by \newcite{hofmann2020appraisal} was developed in an experimental setting, applications of appraisal models on non-experimental data are still missing.  

In this work, we introduce APPReddit, the first corpus of social media posts annotated for appraisal. $1,091$ events gathered from Reddit\footnote{\url{https://www.reddit.com}} have been annotated based on Roseman's model of appraisal~\cite{roseman1991appraisal,roseman2013appraisal}. Far from being the only eligible approach for such a task, Roseman's proposal has been adopted because it provides a comprehensive model of both appraisal and coping strategies, which fits the need of providing a data set focused on the causes that elicit the emotions expressed in a message, rather than the identification of emotions themselves. 

The paper is structured around three research questions.

\smallskip
RQ1: can texts produced in a non-experimental setting be understood and annotated according to Roseman's appraisal model?

\smallskip
RQ2: is it possible to map an annotation scheme based on Rosemans's appraisal theory onto enISEAR, which is based on a different appraisal theory?

\smallskip
RQ3: Can non-experimental and experimental data complement each other towards better computational modeling of appraisal?

The plan of the paper is the following. After a review of existing computational resources for emotion detection (Section \ref{related}), we present the annotation scheme and the most challenging aspects of the task (Section \ref{schema}). In Section \ref{corpus}, results of the annotations are illustrated, with a  focus on how appraisal dimensions correlate in posts. Finally, an experimental mapping of annotation schemes and data is provided (Section \ref{alignment}) and discussed in Section \ref{discussion}.

\section{Related Work} \label{related}

Two main lines of research are relevant to the discussion of our proposal: on the one side, the research in language resources for emotion detection; on the other side, the research in psychological models of emotional appraisal.

\subsection{Resources for Emotion Detection}
The number of corpora, computational resources, and models developed for emotion recognition is wide, and several surveys address different aspects of the topic. 

\newcite{oberlander2018analysis} analyzed $14$ annotated data sets, classifying them along $5$ axes: granularity of annotation (e.g., headlines, tweets, sentences), emotion model that inspired the annotation scheme, size of the corpus, and topic. According to their findings, Ekman's model \cite{ekman1992there} is the most adopted, with $9$ out of $14$ annotation schemes based on the $6$ basic emotions in Ekman's model: joy, anger, sadness, disgust, fear, surprise. The authors mapped all corpora to this model and evaluated how well a 
bags-of-words based classifier trained on a data set predicts emotions expressed in the others, in order to provide a baseline for transfer learning experiments.

\newcite{alswaidan2020survey} reviewed several existing models for emotion recognition, outlining some crucial issues: the identification of implicit expression of emotions; the scarcity of non-English corpora; the lack of large enough data sets with a balanced distribution of emotions.

Another family of computational resources for emotion recognition comprehends event-based annotated data sets. The ISEAR\footnote{International Survey of Emotional Antecedents
and Reactions, \url{https://www.unige.ch/cisa/research/materials-and-online-research/research-material/}} corpus \cite{scherer1994evidence} is the outcome of a cross-cultural psychological study on emotional response. $2,921$ people from $37$ countries were asked to provide a free verbal description of an autobiographical situation related to $7$ emotions: joy, anger, fear, sadness, disgust, shame, and guilt. The survey led to the creation of  a corpus of $7,665$ events. Even if it was not specifically developed for emotion recognition, the ISEAR corpus has been very influential in this field. The Emotinet Ontology \citelanguageresource{balahur2011building}, designed for collecting semantically encoded events, their emotions, and appraisal, was extended with examples from this psychological survey. Appraisal dimensions and their interaction in eliciting emotions were not modeled in ontology, though.
\citelanguageresource{troiano2019crowdsourcing} delivered two corpora of German and English event descriptions for emotion recognition that rely on ISEAR methodology: deISEAR, and enISEAR. Corpora were crowdsourced on Figure-Eight in two rounds of annotations. A first group of annotators generated emotion-focused events of the form `I feel ... when ..'. A second independent group annotated the emotion expressed by events, therefore validating the generated texts.  \newcite{hofmann2020appraisal} completed this work by adding an annotation of $7$ appraisal dimensions on the English corpus: Attention, Certainty, Effort, Pleasantness, Responsibility, Control, and Circumstances. \newcite{hofmann-etal-2021-emotion} later tested different strategies for increasing inter-annotator agreement on these dimensions.

Finally, it is worth mentioning the work of \newcite{ding2018event}, who classified positive and negative affective events according to a taxonomy of $7$ human needs: physiological, health, leisure, social, financial, cognition, and freedom. 

Existing literature shows a limited number of studies investigating the interaction between events and emotion elicitation at the linguistic levels, i.e., resources that could improve models for implicit emotion recognition. Among them, corpora annotated for appraisal on social media data do not exist. APPReddit fills this gap with a corpus of events gathered from social media and annotated for appraisal. To our knowledge, APPReddit is the first resource adopting this theory of emotional appraisal to non-experimental data. 

%APPReddit is an event-based corpus in which the role of appraisal rather than emotion is emphasized. In fact, it does not rely on already emotion-annotated events, but specifically test an appraisal theory not yet adopted in existing corpora. Furthermore, its development is based on social media post, rather than experimental data. 

\subsection{Appraisal Theories}
%FORSE SPOSTARE SU% Approached scientifically for the first time by Charles Darwin \cite{darwin2015expression}, the topic of emotions has been extensively discussed across research domains. Many theories have been proposed that describe emotions and their neural, behavioral and bodily correlates. 

\begin{table*}[ht]
\begin{center}
\begin{tabular}{l|l|l|l|l|l|l|l}

          \textbf{Emotion}&\textbf{Family}&\textbf{Unexpectedness} & \textbf{Certainty} & \textbf{Control}&\textbf{Consistency}&\textbf{Responsibility}&\textbf{Appetitive}\\
     \hline
     \hline
     Hope & Contacting & $-$ & $-$&$-$&$+$&$NA$&$NA$ \\
     Joy & Contacting & $NA$ & $+$&$+$&$+$&$NA$&$+$ \\
     Fear & Distancing & $-$ & $-$&$-$&$-$&$NA$&$-$ \\
     Distress & Distancing & $NA$ & $+$&$-$&$-$&$NA$&$-$ \\
     Regret & Distancing & $NA$ & $+/-$&$-$&$-$&$Self$&$-$ \\
     Anger & Attack & $NA$ & $+/-$&$+$&$-$&$Other$&$-$ \\
     Guilt & Attack & $NA$ & $+/-$&$+$&$-$&$Self$&$-$ \\
     Shame & Rejection & $NA$ & $+/-$&$-$&$-$&$Other$&$-$ \\
     Surprise & $NA$ & + & $NA$ & $NA$ & $NA$ & $Circumstance$ & $NA$ \\

\end{tabular}
 \caption{Examples of interaction between emotions and appraisal according to \protect\newcite{roseman2013appraisal}}
 \label{table:roseman_model}
\end{center}
\end{table*}

In this section, we briefly describe emotion theories and, more specifically, those that focus on the appraisal processes leading to the emotional experiences and to the corresponding coping strategies. 
In affective sciences, the term \textit{appraisal theories}, proposed by \newcite{arnold1960emotion}, refers to a family of theories describing emotions as adaptive responses which reflect cognitive evaluations of features of the environment that are significant for the organism’s well-being~\cite{moors2013appraisal}. Appraisal theories see emotions as processes rather than discrete states and focus on the components of these processes, describing the key components of emotion elicitation, intensity and differentiation, i.e., the emotion that follows the cognitive evaluation of a given event and its intensity \cite{moors2009theories}. 
The term \textit{appraisal} refers to a spontaneous and effortless assessment of the environment against a set of features, named \textit{appraisal variables}; this assessment, along with changes in action tendencies, behavioral responses and bodily reactions, creates an emotional episode. Theories have identified a core set of appraisal variables, such as \textit{goal relevance}, \textit{goal congruence} or \textit{motive consistency}, \textit{certainty}, \textit{coping potential} or \textit{control}, \textit{agency}, and \textit{unexpectedness} \cite{sander2018appraisal}.

In agent theories, the interest for appraisal theories, motivated by the goal of creating believable virtual agents, has led to the integration of appraisal models into agents architectures. 
In particular,~\newcite{marsella2009ema} proposed a general framework for  emotional appraisal and coping in agents, where these two processes interact continuously: their framework, called EMA, has affected several research areas within affective sciences, 
ranging from social robotics~\cite{breazeal2016social} to computational linguistics~\cite{clavel2015sentiment}.
\newcite{dias2014fatima} integrated the appraisal model proposed by~\newcite{ortony1990cognitive} into a virtual agent architecture where the emotional appraisal affects the agent's deliberation and planning to yield a more natural behavior.
%than it would be granted by mere rationality. 

A thorough review of appraisal theories goes beyond the scope of this paper (see: \newcite{dalgleish2000handbook}, \newcite{scherer2001appraisal}, \newcite{moors2013appraisal}). Nevertheless, the appraisal theory proposed by \newcite{roseman1991appraisal} is crucial to understand our work. Roseman's \textit{Emotion System model} \cite{roseman2013appraisal} provides a detailed description of emotions and corresponding appraisal processes, in terms of the different dimensions that are leveraged to evaluate the environment. The model also describes how these  dimensions interact to elicit a given emotion, and the coping responses that follow a given appraisal and the elicited emotion. Table \ref{table:roseman_model} shows a subset of emotions and appraisal dimensions according to \newcite{roseman2013appraisal}. As it can be seen, each emotion is grouped in a family of behaviors --- contacting, distancing, attack, rejection --- and is the product of a specific combination of appraisal dimensions. For instance, \textit{anger} is part of the \textit{Attack} emotion family group and is the product of \textit{high control}, \textit{low consistency}, and \textit{external cause }of the event. 
Despite the availability of appraisal theories, 
Roseman's modeling of emotional responses fits our need for developing a linguisitic resource focused on how emotions, events, and behaviour interact and are explained. Thereby, we chose this specific theory of appraisal to design our annotation scheme.  %Our annotation schema leverages this model to annotate appraisal variables in a set of social media posts.  

\section{Corpus Creation} \label{schema}

APPReddit is a corpus of $500$ Reddit posts annotated for appraisal. 
Each post contains one or more events (for an overall $1,091$ events) annotated on five appraisal dimensions derived from Roseman's model: Certainty, Consistency, Control, Unexpectedness, and Responsibility. 

\begin{figure*}
    \centering
    \includegraphics[width=.9\textwidth]{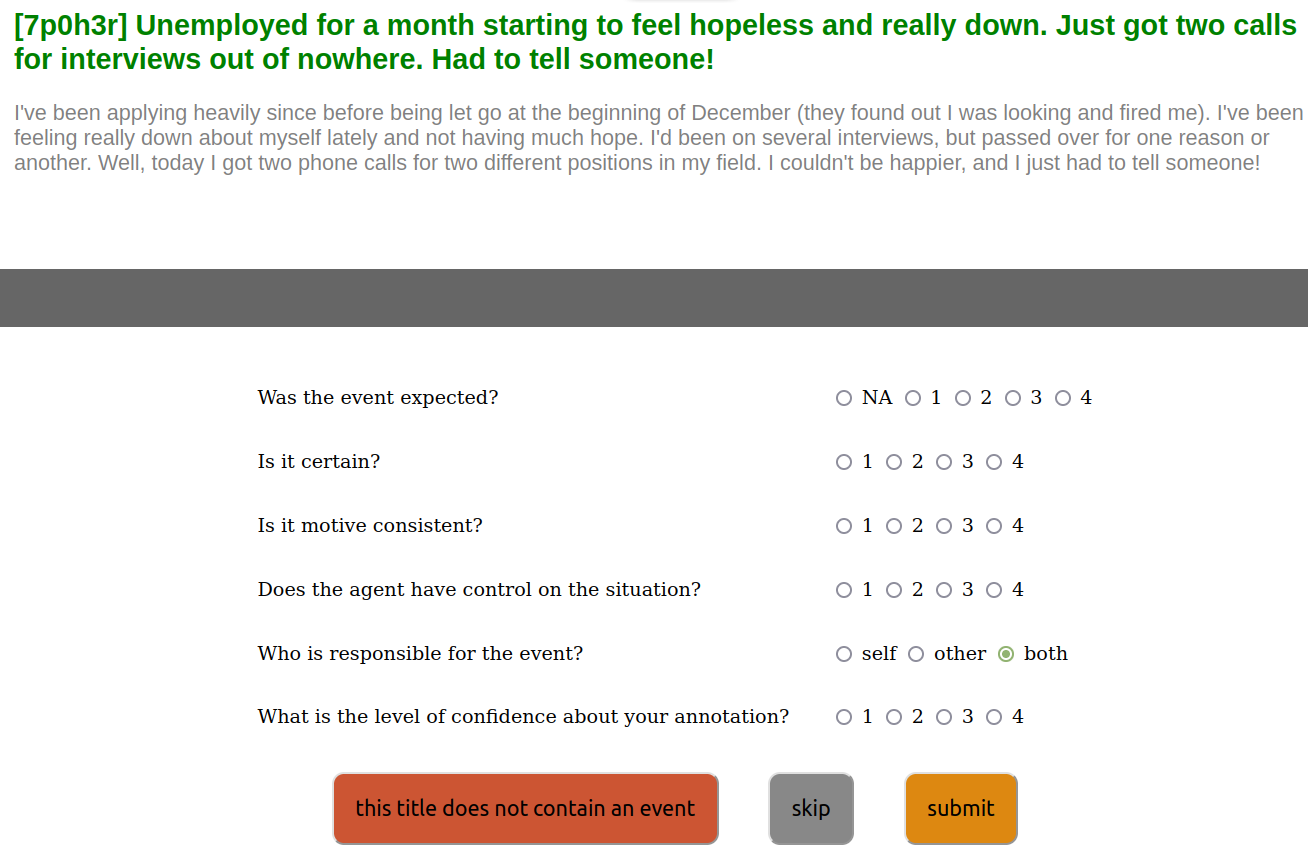}
    \caption{Screenshot of the annotation interface showing the annotation of the title of a post.}
    \label{fig:interface}
\end{figure*}

\subsection{Annotation Scheme}

The annotation scheme was developed based on $5$ out of $7$ appraisal dimensions formulated by Roseman (see Table \ref{table:roseman_appraisal}). 

\textbf{Unexpectedness} measures the extent to which an event took the agent unaware, and correlates with surprise. Highly unexpected events such as the sudden death of a relative may determine high unexpectedness.
\\
\textbf{Consistency} evaluates whether a situation matches agent's goals. Being stuck in a traffic jam is perceived as motive inconsistent by a person who is trying to reach their workplace in time.
\\
\textbf{Certainty} measures the degree of certainty of an event. Having a job interview scheduled for tomorrow can lead to high uncertainty about getting the job, but even events in the past may be uncertain. A low confident student may be not certain about the result of their test.
\\
\textbf{Control} plays a role in the evaluation of how much an agent has control on a situation. Most of the people are likely to have low control on macroeconomic events, while they could have an impact on everyday situations. 
\\
\textbf{Responsibility} is about who is perceived as responsible for a situation. The author of a message about being harassed by their boss is identifying the cause of a situation as someone/something external. A person who is telling their story about how they overcome depression may consider themselves responsible for this situation. 

\smallskip
The following two dimensions of Roseman's Emotion System Model~\cite{roseman2013appraisal} are not included in our annotation scheme.

\textbf{Motivational State} is not part of the annotation schema for two reasons. This dimension is the most correlated with coping strategies, since it focuses on distancing or contacting with an event that may be perceived as punishing or rewarding by the agent. During a first trial annotation, this appraisal dimension seemed to overlap with coping strategies adopted by the user who posted the message. Furthermore, this kind of evaluation often appears to be pragmatic, thus not having an explicit manifestation in the text. 

\textbf{Problem Type} focuses on whether a situation is intrinsically motive-inconsistent or not. We did not consider it as part of the annotation schema because it could be inferred directly from the Consistency dimension.

Except for Responsibility, each dimension in our annotation schema is evaluated on a scale from $1$ (very low presence) to $4$ (very high presence). Agency is the only dimension with a nominal set of options: \textit{self}, \textit{other}, and \textit{both}. 

The annotation of Unexpectedness had to be evaluated as `Not Applicable' for each event not yet happened. Remaining dimensions could be marked as `Irrelevant' only together.

\begin{table}[ht]
\begin{center} 
\begin{tabularx}{\columnwidth}{l|X}

      \textbf{Appraisal Dimension} & \textbf{Annotation Scheme}\\
      \hline
      \hline
      Unexpectedness & Is the event expected?\\
      Consistency & Is the event motive consistent?\\
     Certainty & Is the event certain?\\
     Control & Does the user have control over the situation?\\
   Responsibility & Who is responsible for the event?\\
  %   Motivational State & X \\
  %   Problem Type & X\\

\end{tabularx}
\caption{Roseman's model of appraisal, adapted from \protect\newcite{roseman2001model}, mapped onto the annotation scheme.}
\label{table:roseman_appraisal}
 \end{center}
\end{table}
\subsection{Data Collection and Annotation}
Selecting non-experimental data to be annotated for appraisal is not a trivial operation. The texts shall express one or more situations or events triggering an emotion, and they have to be long enough for such a fine-grained annotation. Reddit responds to this need because instead of being a continuous stream of content, like other social media, its structure is similar to a collection of forums. It is organized in thematic channels (called \textit{subreddits}), where users can start threads about a disparate range of topics, including sharing their personal issues and emotions. After a review of the public subreddits, we selected $20$ of them. The full list of subreddits is the following: Anger,
offmychest,
helpmecope,
anxiety,
mentalhealth,
relationship\_advice,
rant,
DecidingToBeBetter,
CasualConversation,
getting\_over\_it,
UnsentLetters,
apologizeplease,
changemyview,
DearPeople,
Dear\_Ex,
dearsincerely,
TrueOffMyChest,
DiaryOfARedditor,
MMFB,
confessions. From this pool, we gathered all posts containing at least $5$ sentences. We then selected a random sample of $500$ posts to be tested over the annotation scheme. For the purpose of this data collection, we only considered the textual posts starting new threads including their titles, ignoring the following comments by other users, which could otherwise only be interpreted in the context of the full thread.

A first round of pilot annotation over a small amount of posts revealed a mismatch between the task and the corpus. If, on the one hand, annotators often perceived multiple events or situations as present in posts, on the other hand, an evaluation of the appraisal at the sentence level was problematic because many sentences did not reference emotional content. Thereby, we reviewed the data set and manually grouped sentences in coherent subgroups of consecutive or nonconsecutive sentences, corresponding to events. Following cognitive theories on Event Segmentation \cite{zacks2007event}, we use the term \textit{event} to indicate a portion of time ``perceived by an observer as having a beginning and an end'', and featuring no substantial changes in situational dimensions such as main characters, goals or interaction among characters \cite{zacks2009segmentation}. We followed two criteria: sentences which shared the same location, time, and participants were grouped in single events (Cfr: \newcite{alrashid2021pilot}); external diegetic events (e.g., the telling of an accident) were distinguished from internal extradiegetic events (e.g., the manifestation of a certain emotional state), following the work of \newcite{swanson2017empirical} where this distinction was introduced to keep reported events and comments separated in the annotation of personal events.

For example, the post below (example $e_1$) was divided in three sub-events. The first is a situation, staged in the past, and the user is the agent. The second is a recent fact, characterized by having a different agent. Finally, the third event is located in the past, and it is extradiegetic, since the author explicitly express her personal guilt and regrets rather than telling a story. Titles were always considered as single events.

\smallskip
($e_1$) \textit{$<$EVENT\_1$>$I used to babysit for my neighbors two children about 4 years ago.$<$/EVENT\_1$>$ $<$EVENT\_2$>$TIL that one of them OD'd on pills Friday night and died two days later this Sunday. He was 16 and I feel horrible.$<$/EVENT\_2$>$ $<$EVENT\_3$>$I feel that somehow this is my fault, and that I didn't spend enough time playing with them, or playing the right games. If I were a better babysitter, I should have made some sort of lasting impression on these kids and this wouldn't have happened.$<$/EVENT\_3$>$}.

\smallskip
The three events forming example ($e_1$) express different appraisal configurations, as it resulted from the annotation process. 
\begin{itemize}
    \item EVENT\_1 was evaluated as highly Certain (4), with low Control (1) and Consistency (1), and the responsibility was attributed to the agent (self). Unexpectedness was marked as Not Applicable.
    \item Certainty (4), Control (1), and Consistency (1) were annotated the with the same values in EVENT\_2, which differs from the former in being highly unexpected (1), and having an external responsible (other).
    \item EVENT\_3 was evaluated as an event in which the author had high Control (3), and she was responsible for it (self). Certainty (4), Unexpectedness (1), and Consistency (4) were annotated the same as EVENT\_2.
\end{itemize}

After the event identification stage, the data set included $1,091$ events. We collected 2 annotation of appraisal for each event from 5 annotators: 2 males, and 3 females; 3 PhD students, 1 university teacher, and a post Doc. Annotators were trained with a meeting in which guidelines were presented. Each annotated the same $10$ posts and most difficult cases were reviewed together, in order to reduce the impact of subjectivity within the task.
For the annotation task, a graphical interface was created (Figure \ref{fig:interface}). The upper part contains the title and the entire post with the event to annotated highlighted in green. In the lower part the appraisal dimensions can be annotated on a 4-point scale with the exception of Responsibility, for which three categorical dimensions are available. A self-assessment entry and a comment box (not visible in the figure) are posited below the appraisal annotation.

The inter-annotator agreement was calculated with Krippendroff's Alpha\footnote{This metric fit our case, since annotations were sparse.} and is generally low, ranging from $0.38$ (Unexpectedness) to $0.48$ (Consistency), as shown in Table \ref{table:krippendorff}. These values are comparable to those obtained by \newcite{hofmann2020appraisal} on experimental data, although the latter were computed with a different metric (Cohen's kappa): in enISEAR, in fact, the average inter-annotator agreement is $0.53$, with a significant variation from a minimum of $0.31$ to a maximum of $0.89$. 
More precisely, in APPReddit, the agreement is lower for Responsibility (alpha = $0.41$ vs. averaged kappa $0.68$) and Consistency (alpha = $0.48$ vs. averaged kappa $0.89$). Conversely, Certainty is characterized by a higher inter-annotator agreement in APPReddit (alpha = $0.44$ vs. averaged kappa $0.33$). In general, a lower agreement was expected due to the nature of Reddit data, which are spontaneous and therefore often more difficult to interpret.
For example, many messages with sentences such as ``Please make me feel better'' convey a pragmatic function in addition to the event. This may lead to divergent interpretations of what constitutes the main event: some annotators identified the event in the act of asking for help, whereas others focused on the event which the user wishes to happen (namely, feeling better).

Other agreement issues were dimension-specific. For instance, the following event ($e_2$) has been interpreted as highly consistent ($3$) by the first annotator for its reference to people with ``more catastrophic problems'' than the appraising agent (thus, minimizing the mismatch with the agent's goals, according to the definition of Consistency); conversely, the second annotator interpreted the event as poorly Consistent ($1$) focusing on the self-pity expressed by the author about people telling them ``to be grateful'' (and thus emphasising the mismatch with the agent's goal).

\smallskip
($e_2$) \textit{and if I was rich I would donate 90\% of my funds to helping those people, but just because people have more catastrophic problems than me, doesn't mean I don't have a right to cry, yell, complain, etc. And I just get tired of people telling me to be grateful, everyone has problems, we have a right to complain without taking in to factor other world problems, please kill yourself.}

\begin{table}[ht!]
\begin{center}
\begin{tabularx}{\columnwidth}{l|X}

      \textbf{Appraisal Dimension} & \textbf{Krippendorff's Alpha}\\
      \hline
      \hline
      Unexpectedness & $0.36$\\
      Consistency & $0.48$\\
     Certainty & $0.44$\\
     Control & $0.38$\\
     Responsibility & $0.41$\\

\end{tabularx}
\caption{The Krippendroff's Alpha score for each appraisal dimension that was annotated.}
 \label{table:krippendorff}
 \end{center}
\end{table}

In order to account for this type of divergences, we defined two types of agreement: `agreement' when the two annotators labeled the event with the same scalar value; `partial agreement' for the cases where annotators' values different by one point in the scale (e.g., 2 and 3) and calculated the mean. All the remaining events were labeled by a third annotator, who solved the disagreement. For instance, Consistency in $e_2$ was marked as low (1). Finally, we mapped scalar values to dichotomous nominal categories: appraisal dimensions with a score equal or lower than $2$ were mapped to $0$ (low) while dimensions with a score above $2$ were mapped to $1$ (high). \\The annotation scheme for Responsibility was nominal, therefore there were no `partial agreement' cases. When mapped to dichotomous categories, we merged `both' and `self' responsibility values into $1$ (high: events in which the user is totally or partially responsible) while `other' was mapped to $0$ (low: events where the user is not responsible). 

\section{Corpus Description} \label{corpus}

\begin{figure}[ht]
\begin{center}
%\fbox{\parbox{6cm}{
%This is a figure with a caption.}}
% old picture \includegraphics[scale=0.5]{lrec2020W-image1.eps} 
\includegraphics[width=0.5\textwidth]{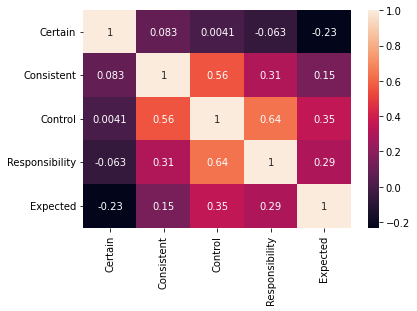} 
\caption{Correlation of appraisal dimensions within the APPReddit corpus (Spearman's $\rho$).}
\label{img:correlation}
\end{center}
\end{figure}

A first overview of the corpus (see Table \ref{table:labels_distribution}) shows a moderate imbalance towards low-labelled dimensions for Unexpectedness ($0.43$ low vs. $0.28$ high), Consistency ($0.53$ low vs. $0.36$ high), and Control ($0.54$ low vs. $0.35$ high). In enISEAR, both Consistency ($0.85$ low vs. $0.15$ high) and Control ($0.78$ low vs. $0.22$ high) are also skewed toward low values, but with a stronger imbalance. Certainty ($0.10$ low vs. $0.79$ high) was instead mostly annotated as high in our corpus, similarly and actually more strongly unbalanced than enISEAR ($0.24$ low vs. $0.76$ high). This seems to be due to the type of data we annotated. In fact, Reddit posts often report events and situations happened in the past, that are therefore more likely to be certain. Responsibility in APPReddit ($0.4$ low vs. $0.5$ high) is annotated with an inverse trend with respect to esISEAR ($0.62$ low vs. $0.38$ high), possibly due to the autobiographical nature of the posts.
Finally, a relevant tendency appears to be the high number of events where Unexpectedness is not applicable ($0.29$ of the total), since all events that are not yet happened do not have a value for this dimension.

\begin{table}[ht]
\begin{center}\small
\begin{tabularx}{\columnwidth}{l|X|X|X}

      \textbf{APPReddit corpus} & \textbf{Low} & \textbf{High} & \textbf{NA}\\
      \hline
      \hline
      Unexpectedness & $0.43$ & $0.28$ & $0.29$\\
      Consistency & $0.53$ & $0.36$ & $0.11$\\
     Certainty & $0.10$ & $0.79$ & $0.11$\\
     Control & $0.54$ & $0.35$ & $0.11$\\
     Responsibility & $0.40$ & $0.50$ & $0.10$\\
     \hline
     \textbf{enISEAR corpus} & \textbf{Low} & \textbf{High} & \textbf{NA}\\
      \hline
      \hline
      Certainty & $0.24$ & $0.76$ & $0$\\
      Consistency & $0.85$ & $0.15$ & $0$\\
      Responsibility & $0.62$ & $0.38$ & $0$\\
      Control & $0.78$ & $0.22$ & $0$\\
      Attention & $0.33$ & $0.67$ & $0$\\
      Effort & $0.60$ & $0.40$ & $0$\\
      Circumstance & $0.76$ & $0.24$ & $0$\\

\end{tabularx}
\caption{The percentage distribution of labels in APPReddit and enISEAR corpora (with labels mapped from a 4-value scale to two dichotomous nominal categories).}
\label{table:labels_distribution}
 \end{center}
\end{table}

Given the multi-dimensional annotation scheme of appraisal dimensions, we computed Spearman's rank correlation to evaluate the correlation between them pairwise. Results (Figure~\ref{img:correlation}) show interesting correlations between appraisal dimensions. Control strongly correlates with both Responsibility ($0.64$), and Consistency ($0.56$) which in turn shows a moderate correlation with Responsibility ($0.31$). Hence, events and situations may be consistent when the agent has control on them, and they are responsible for their happening, as in $e_3$ (below). 

\smallskip
($e_3$) \textit{Go from ``clean my house'' to ``stand up from the couch, put the dish beside the sink, fill up the sink with water''. Its been working better for me, even if I feel a little silly with my teeny tiny steps.}

In a complementary way, negative events seem to be affected by a lack of control and by an external responsibility. Event ($e_4$) exemplifies such a correlation:

($e_4$) \textit{Monday I was flying to Idaho because my mother was suicidal due to finding out her boyfriend has severe liver cancer.}

\smallskip
Unexpectedness shows a significant inverse correlation with Certainty ($-0.23$). Intuitively, a low expected event is likely to be highly certain since it is already happened. For instance, in example $e_5$, the user expresses surprise for a letter they unexpectedly received, and this event is considered certain. 

\smallskip
($e_5$) \textit{It has been 5 days since I received letters from the person who I have been so angry with, and it's proving to be very effective. I really wanted to say thank you because I don't know how much longer I would have been plagued by my anger and anxiety.}

\section{Appraisal models and corpora alignment} \label{alignment}
\begin{table*}[ht]
\begin{center}
\small
\begin{tabular}{l|l|c|c|c|c}

         Training set & Test set &Certainty&Consistency&Responsibility&Control\\
     \hline
     \hline
     APPReddit & APPReddit & $0.832$ & $0.675$ & $0.688$ & $0.507$ \\
     enISEAR & APPReddit & $0.844$ & $0.450$ & $0.318$ & $0.455$ \\
     \hline
     enISEAR & enISEAR & $0.684$ & $0.840$ & $0.616$ & $0.685$ \\
     APPReddit & enISEAR & $0.651$ & $0.841$ & $0.551$ & $0.699$ \\
     \hline
     APPReddit+enISEAR & enISEAR & $0.674$ & $0.870$ & $0.658$ & $0.712$ \\
     APPReddit+enISEAR & APPReddit & $0.832$ & $0.646$ & $0.689$ & $0.510$ \\

\end{tabular}
 \caption{Results of mapping experiments between APPReddit and enISEAR corpus in terms F1-scores. }
 \label{table:experiments}
\end{center}
\end{table*}

Since APPReddit and enISEAR differ in the annotation scheme and the kind of annotated data, this section is devoted to evaluate whether the two resources can be mapped onto each other for joint use in the creation of analysis tools. With such a mapping, a larger, more balanced, and multi-domain corpus annotated for appraisal would be available.

The APPReddit annotation scheme was derived from \newcite{roseman1991appraisal} whereas enISEAR's relies on \newcite{smith1985patterns}. The former includes $5$ dimensions: Unexpectedness, Certainty, Control, Responsibility, and Consistency. The latter includes 7 dimensions: Certainty, Control, Responsibility, Pleasantness, Anticipation, Effort, and Circumstance. Anticipation and Effort were not mapped because they are considered coping strategies by Roseman (the former related to Hope, the latter to Distress). Unexpectedness is related to Surprise in \newcite{roseman1991appraisal}, while \newcite{smith1985patterns} do not include this dimension, considering the emotion elicitation as the joint presence of low responsibility and control. Circumstance and Responsibility differs in assigning the responsibility of an event to an agent, which could be self or another, or other causes not attributable to living people. This distinction is present in both theories, but does not result in two separate appraisal dimensions in Roseman's \cite{roseman2013appraisal}.  

Given these differences, we decided to limit the mapping to the $4$ dimensions that appear likewise in annotation schemes and appraisal models: Certainty, Control, Responsibility, and Consistency/Pleasantness. We then mapped the two ordinal scales (0-3 adopted for enISEAR; 1-4 for APPReddit) to nominal dichotomous categories. Scores lower or equal than $1$ in enISEAR and lower or equal than $2$ in APPReddit were mapped to $0$ (low); the other scores were mapped to $1$ (high).  

\smallskip
After normalizing the two corpora, we performed three experiment of appraisal prediction.

\begin{enumerate}
    \item Predicting appraisal dimensions of APPReddit using enISEAR as training set
    \item Predicting appraisal dimensions of enISEAR using APPReddit as training set
    \item Evaluating performances of a concatenation of the two corpora
\end{enumerate}

For the experiment, we implemented a binary classifier based on Support Vector Machine (SVM) rather than a SOTA transformer model in order to compare only linguistic information provided by the two data sets. 

In particular, the SVM classifier is employed with the radial basis function kernel (RBF) using the default parameters of $C$ and $\gamma$ provided by scikit-learn library \cite{scikit-learn}.
As input, we used a simple bag-of-words representation, extracting from the texts the unigrams of words and weighting them with TF-IDF (Term Frequency–Inverse Document Frequency) measure. The tokenization of the texts is performed by the function of vectorization provided by scikit-learn. 

After splitting the two corpora in a training set ($80$\% of the data) and a test set ($20$\% of the data), we first trained the classifier for evaluating internal consistency. Then we trained it with a corpus to predict the other's test set. Finally we trained a concatenation of the two set and predicted both APPReddit and enISEAR test sets.

%We trained the classifier with the enISEAR training set, APPReddit, and finally a concatenation of the two sets.
The results in Table \ref{table:experiments} show that prediction of APPReddit with enISEAR training set led to a significant drop of F1-score in predicting Consistency ($0.45$ vs $0.67$), Responsibility ($0.31$ vs $0.68$), and Control ($0.50$ vs $0.45$), while there is an improvement in the prediction of Certainty ($0.84$ vs $0.83$). \\
APPReddit predicts well enISEAR Consistency and Control with an F1-score of $0.841$ and $0.699$. There is however a drop in predicting Responsibility ($0.55$ vs $0.61$) and Certainty ($0.65$ vs $0.68$).

In both the experimental setups the drop of performances seems to be consistent with the observations in Section \ref{corpus} about its different distribution in the two data sets. The choice of not mapping the Circumstance appraisal dimension (non-agent responsibility of an event) may has had a role in such a drop of the F1-score. 
The last experiment, which relied on merging both corpora in a unique training set, achieved encouraging performances. The concatenation of APPReddit and enISEAR showed equal or better performances on Consistency, Responsibility, and Control with enISEAR as test set, and a limited drop in predicting Certainty ($0.67$ vs $0.68$). It also predicted well on three dimensions from the APPReddit test set: Certainty, Responsibility and Control, with a little loss of performance for Consistency ($0.64$ vs $0.67$). It is worth mentioning the impact of such merging on Responsibility if compared to other experiments: enISEAR alone showed a F1-score drop of $0.37$ point in predicting this dimension in the APPReddit test set, while the concatenation led to an increase of $0.001$. Similarly, the merged corpora increased the F1-score by $0.10$ in predicting Responsibility from enISEAR test set, if compared to APPReddit alone. Such results are encouraging if compared to existing evaluations of the alignment between corpora for emotion recognition \cite{oberlander2018analysis}.
%
%The merged training sets also compensates for the drop on Certainty prediction with the APPReddit training set, which is due to high imbalance of the corpus on that dimension.
 
The good alignment shows that a set of appraisal dimensions seem to occur consistently in different types of data.
%, namely, natural data and experimentally collected data from enISEAR. 
This paves the way to transferring this knowledge to other domains, such as abusive language detection and stance detection: these phenomena, in fact, could be better explained in the light of different appraisal configurations.      

\section{Discussion}\label{discussion}
In this work, we presented a novel corpus of social media data annotated for appraisal. The corpus was aligned with enISEAR and an experiment to evaluate the mapping was performed.
The results of the annotation and the mapping experiments provide answers to our three research questions.

\smallskip
\textbf{RQ1: can texts produced in a non-experimental setting be understood and annotated according to Roseman's appraisal model?}

An overall analysis of the corpus shows that Roseman's appraisal model can be applied to texts collected in a non-experimental setting. Apart from Certainty, all appraisal dimensions are moderately balanced. Many interesting correlations between them emerged, namely Control and Consistency, Control and Responsibility, and Unexpectedness and Certainty. This suggests expanding the corpus with texts from other domains. Furthermore, an annotation of emotion types and coping strategies could be useful to better understand the relationship between the characteristics of events and the emotion types.

\smallskip
\textbf{RQ2: is it possible to map an annotation scheme based on Rosemans's appraisal theory to enISEAR, which is modeled on a different appraisal theory?} 

Despite the differences between APPReddit and enISEAR annotation schemes, $4$ appraisal dimensions are common to both and can be mapped. This partially reduces the unbalance in the two corpora, especially regarding Certainty for APPReddit  and Pleasantness for enISEAR. A further step may be the application to the corpus of existing resources for emotion detection, in order to enrich the resource with information about the emotions correlating with the appraisal dimensions.

\smallskip
\textbf{RQ3: Can non-experimental and experimental data complement each other towards better computational modeling of appraisal?} 
Experiments showed a good performance in predicting appraisal dimensions between the two  corpora. This confirms the quality of mapping. Further transfer learning experiments will validate whether this knowledge can be leveraged in other tasks, such as abusive language and stance detection.

\section{Conclusion and Future Work} 

In this paper we introduced APPReddit, a novel corpus of $1091$ events gathered from social media and annotated for appraisal. The corpus was aligned with an existing resource of events collected in an experimental setting and annotated for appraisal with a different annotation scheme. Results showed consistency between the two corpora despite they include different types of data.

Future work will be devoted to expanding the corpus quantitatively, including messages from other sources. Furthermore, the annotation scheme will be improved to integrate the identification of coping strategies and emotion types. 

Finally, transfer learning experiments will be performed in order to test the effectiveness of this resource in other domains where emotional responses to events may improve prediction of sentiment and the explainability of NLP models.

\section*{Acknowledgements}

The work of S. Frenda and V. Patti was funded by the research projects ``STudying European Racial Hoaxes and sterEOTYPES'' (STERHEOTYPES, under the call ``Challenges for Europe'' of VolksWagen Stiftung and Compagnia di San Paolo). 
The work of V. Basile was funded by the project ``Be Positive!'' (under the 2019 ``Google.org Impact Challenge on Safety'' call).

\section{Bibliographical References}\label{reference}
\bibliographystyle{lrec2022-bib}
\bibliography{appraisal}

\begin{thebibliography}{}

\bibitem[\protect\citename{Alrashid and Gaizauskas}2021]{alrashid2021pilot}
Alrashid, T. and Gaizauskas, R.~J.
\newblock (2021).
\newblock A pilot study on annotating scenes in narrative text using sceneml.
\newblock In {\em Text2Story@ ECIR}, pages 7--14.

\bibitem[\protect\citename{Alswaidan and Menai}2020]{alswaidan2020survey}
Alswaidan, N. and Menai, M. E.~B.
\newblock (2020).
\newblock A survey of state-of-the-art approaches for emotion recognition in
  text.
\newblock {\em Knowledge \& Information Systems}, 62(8).

\bibitem[\protect\citename{Arnold}1960]{arnold1960emotion}
Arnold, M.~B.
\newblock (1960).
\newblock {\em Emotion and personality.}
\newblock Columbia University Press.

\bibitem[\protect\citename{Bradley and Lang}2007]{bradley2007emotion}
Bradley, M.~M. and Lang, P.~J., (2007).
\newblock {\em Emotion and motivation.}, page 581–607.
\newblock Cambridge University Press.

\bibitem[\protect\citename{Breazeal \bgroup et al.\egroup
  }2016]{breazeal2016social}
Breazeal, C., Dautenhahn, K., and Kanda, T.
\newblock (2016).
\newblock Social robotics.
\newblock {\em Springer handbook of robotics}, pages 1935--1972.

\bibitem[\protect\citename{Clavel and Callejas}2015]{clavel2015sentiment}
Clavel, C. and Callejas, Z.
\newblock (2015).
\newblock Sentiment analysis: from opinion mining to human-agent interaction.
\newblock {\em IEEE Transactions on affective computing}, 7(1):74--93.

\bibitem[\protect\citename{Dalgleish and Power}2000]{dalgleish2000handbook}
Dalgleish, T. and Power, M.
\newblock (2000).
\newblock {\em Handbook of cognition and emotion}.
\newblock John Wiley \& Sons.

\bibitem[\protect\citename{Dias \bgroup et al.\egroup }2014]{dias2014fatima}
Dias, J., Mascarenhas, S., and Paiva, A.
\newblock (2014).
\newblock Fatima modular: Towards an agent architecture with a generic
  appraisal framework.
\newblock In {\em Emotion modeling}, pages 44--56. Springer.

\bibitem[\protect\citename{Ding \bgroup et al.\egroup }2018]{ding2018event}
Ding, H., Jiang, T., and Riloff, E.
\newblock (2018).
\newblock Why is an event affective? classifying affective events based on
  human needs.
\newblock In {\em Workshops at the Thirty-Second AAAI Conference on Artificial
  Intelligence}.

\bibitem[\protect\citename{Ekman}1992]{ekman1992there}
Ekman, P.
\newblock (1992).
\newblock Are there basic emotions?
\newblock {\em Psychological Review}, 99(3).

\bibitem[\protect\citename{Hofmann \bgroup et al.\egroup
  }2021]{hofmann-etal-2021-emotion}
Hofmann, J., Troiano, E., and Klinger, R.
\newblock (2021).
\newblock Emotion-aware, emotion-agnostic, or automatic: Corpus creation
  strategies to obtain cognitive event appraisal annotations.
\newblock In {\em Proceedings of the Eleventh Workshop on Computational
  Approaches to Subjectivity, Sentiment and Social Media Analysis}, pages
  160--170, Online, April. Association for Computational Linguistics.

\bibitem[\protect\citename{Marsella and Gratch}2009]{marsella2009ema}
Marsella, S.~C. and Gratch, J.
\newblock (2009).
\newblock Ema: A process model of appraisal dynamics.
\newblock {\em Cognitive Systems Research}, 10(1):70--90.

\bibitem[\protect\citename{Moors \bgroup et al.\egroup
  }2013]{moors2013appraisal}
Moors, A., Ellsworth, P.~C., Scherer, K.~R., and Frijda, N.~H.
\newblock (2013).
\newblock Appraisal theories of emotion: State of the art and future
  development.
\newblock {\em Emotion Review}, 5(2):119--124.

\bibitem[\protect\citename{Moors}2009]{moors2009theories}
Moors, A.
\newblock (2009).
\newblock Theories of emotion causation: A review.
\newblock {\em Cognition and emotion}, 23(4):625--662.

\bibitem[\protect\citename{Oberl{\"a}nder and
  Klinger}2018]{oberlander2018analysis}
Oberl{\"a}nder, L. A.~M. and Klinger, R.
\newblock (2018).
\newblock An analysis of annotated corpora for emotion classification in text.
\newblock In {\em Proceedings of the 27th International Conference on
  Computational Linguistics}, pages 2104--2119.

\bibitem[\protect\citename{Ortony \bgroup et al.\egroup
  }1990]{ortony1990cognitive}
Ortony, A., Clore, G.~L., and Collins, A.
\newblock (1990).
\newblock {\em The cognitive structure of emotions}.
\newblock Cambridge university press.

\bibitem[\protect\citename{Pedregosa \bgroup et al.\egroup }2011]{scikit-learn}
Pedregosa, F., Varoquaux, G., Gramfort, A., Michel, V., Thirion, B., Grisel,
  O., Blondel, M., Prettenhofer, P., Weiss, R., Dubourg, V., Vanderplas, J.,
  Passos, A., Cournapeau, D., Brucher, M., Perrot, M., and Duchesnay, E.
\newblock (2011).
\newblock Scikit-learn: Machine learning in {P}ython.
\newblock {\em Journal of Machine Learning Research}, 12:2825--2830.

\bibitem[\protect\citename{Plutchik}1991]{plutchik1991emotions}
Plutchik, R.
\newblock (1991).
\newblock {\em The emotions}.
\newblock University Press of America.

\bibitem[\protect\citename{Roseman and Smith}2001]{roseman2001appraisal}
Roseman, I.~J. and Smith, C.~A.
\newblock (2001).
\newblock Appraisal theory.
\newblock {\em Appraisal processes in emotion: Theory, methods, research},
  pages 3--19.

\bibitem[\protect\citename{Roseman}1991]{roseman1991appraisal}
Roseman, I.~J.
\newblock (1991).
\newblock Appraisal determinants of discrete emotions.
\newblock {\em Cognition \& Emotion}, 5(3):161--200.

\bibitem[\protect\citename{Roseman}2001]{roseman2001model}
Roseman, I.~J.
\newblock (2001).
\newblock A model of appraisal in the emotion system.
\newblock {\em Appraisal processes in emotion: Theory, methods, research},
  pages 68--91.

\bibitem[\protect\citename{Roseman}2013]{roseman2013appraisal}
Roseman, I.~J.
\newblock (2013).
\newblock Appraisal in the emotion system: Coherence in strategies for coping.
\newblock {\em Emotion Review}, 5(2):141--149.

\bibitem[\protect\citename{Russell}1980]{russell1980circumplex}
Russell, J.~A.
\newblock (1980).
\newblock A circumplex model of affect.
\newblock {\em Journal of personality and social psychology}, 39(6):1161.

\bibitem[\protect\citename{Sander \bgroup et al.\egroup
  }2018]{sander2018appraisal}
Sander, D., Grandjean, D., and Scherer, K.~R.
\newblock (2018).
\newblock An appraisal-driven componential approach to the emotional brain.
\newblock {\em Emotion Review}, 10(3):219--231.

\bibitem[\protect\citename{Scherer and Wallbott}1994]{scherer1994evidence}
Scherer, K.~R. and Wallbott, H.~G.
\newblock (1994).
\newblock Evidence for universality and cultural variation of differential
  emotion response patterning.
\newblock {\em Journal of personality and social psychology}, 66(2):310.

\bibitem[\protect\citename{Scherer \bgroup et al.\egroup
  }2001]{scherer2001appraisal}
Scherer, K.~R., Schorr, A., and Johnstone, T.
\newblock (2001).
\newblock {\em Appraisal processes in emotion: Theory, methods, research}.
\newblock Oxford University Press.

\bibitem[\protect\citename{Smith and Ellsworth}1985]{smith1985patterns}
Smith, C.~A. and Ellsworth, P.~C.
\newblock (1985).
\newblock Patterns of cognitive appraisal in emotion.
\newblock {\em Journal of personality and social psychology}, 48(4):813.

\bibitem[\protect\citename{Swanson \bgroup et al.\egroup
  }2017]{swanson2017empirical}
Swanson, R., Gordon, A.~S., Khooshabeh, P., Sagae, K., Huskey, R., Mangus, M.,
  Amir, O., and Weber, R.
\newblock (2017).
\newblock An empirical analysis of subjectivity and narrative levels in weblog
  storytelling across cultures.
\newblock {\em Dialogue \& Discourse}, 8(2):105--128.

\bibitem[\protect\citename{Zacks and Swallow}2007]{zacks2007event}
Zacks, J.~M. and Swallow, K.~M.
\newblock (2007).
\newblock Event segmentation.
\newblock {\em Current directions in psychological science}, 16(2):80--84.

\bibitem[\protect\citename{Zacks \bgroup et al.\egroup
  }2009]{zacks2009segmentation}
Zacks, J.~M., Speer, N.~K., and Reynolds, J.~R.
\newblock (2009).
\newblock Segmentation in reading and film comprehension.
\newblock {\em Journal of Experimental Psychology: General}, 138(2):307.

\end{thebibliography}


\begin{thebibliography}{}

\bibitem[\protect\citename{Balahur \bgroup et al.\egroup
  }2011]{balahur2011building}
Balahur, A., Hermida, J.~M., and Montoyo, A.
\newblock (2011).
\newblock Building and exploiting emotinet, a knowledge base for emotion
  detection based on the appraisal theory model.
\newblock {\em IEEE transactions on affective computing}, 3(1):88--101.

\bibitem[\protect\citename{Bertola and Patti}2016]{bertola2016ontology}
Bertola, F. and Patti, V.
\newblock (2016).
\newblock Ontology-based affective models to organize artworks in the social
  semantic web.
\newblock {\em Information Processing \& Management}, 52(1):139--162.

\bibitem[\protect\citename{Buechel and Hahn}2017]{buechel2017emobank}
Buechel, S. and Hahn, U.
\newblock (2017).
\newblock Emobank: Studying the impact of annotation perspective and
  representation format on dimensional emotion analysis.
\newblock In {\em Proceedings of the 15th Conference of the European Chapter of
  the Association for Computational Linguistics: Volume 2, Short Papers}, pages
  578--585.

\bibitem[\protect\citename{Cambria \bgroup et al.\egroup
  }2010]{cambria2010senticnet}
Cambria, E., Speer, R., Havasi, C., and Hussain, A.
\newblock (2010).
\newblock Senticnet: A publicly available semantic resource for opinion mining.
\newblock In {\em 2010 AAAI fall symposium series}.

\bibitem[\protect\citename{Hofmann \bgroup et al.\egroup
  }2020]{hofmann2020appraisal}
Hofmann, J., Troiano, E., Sassenberg, K., and Klinger, R.
\newblock (2020).
\newblock Appraisal theories for emotion classification in text.
\newblock In {\em Proceedings of the 28th International Conference on
  Computational Linguistics}, pages 125--138.

\bibitem[\protect\citename{Li \bgroup et al.\egroup }2017]{li2017dailydialog}
Li, Y., Su, H., Shen, X., Li, W., Cao, Z., and Niu, S.
\newblock (2017).
\newblock Dailydialog: A manually labelled multi-turn dialogue dataset.
\newblock {\em arXiv preprint arXiv:1710.03957}.

\bibitem[\protect\citename{Mohammad}2018]{mohammad2018obtaining}
Mohammad, S.
\newblock (2018).
\newblock Obtaining reliable human ratings of valence, arousal, and dominance
  for 20,000 english words.
\newblock In {\em Proceedings of the 56th Annual Meeting of the Association for
  Computational Linguistics (Volume 1: Long Papers)}, pages 174--184.

\bibitem[\protect\citename{Strapparava and
  Mihalcea}2007]{strapparava2007semeval}
Strapparava, C. and Mihalcea, R.
\newblock (2007).
\newblock Semeval-2007 task 14: Affective text.
\newblock In {\em Proceedings of the Fourth International Workshop on Semantic
  Evaluations (SemEval-2007)}, pages 70--74.

\bibitem[\protect\citename{Troiano \bgroup et al.\egroup
  }2019]{troiano2019crowdsourcing}
Troiano, E., Pad{\'o}, S., and Klinger, R.
\newblock (2019).
\newblock Crowdsourcing and validating event-focused emotion corpora for german
  and english.
\newblock In {\em Proceedings of the 57th Annual Meeting of the Association for
  Computational Linguistics}, pages 4005--4011.

\bibitem[\protect\citename{Warriner \bgroup et al.\egroup
  }2013]{warriner2013norms}
Warriner, A.~B., Kuperman, V., and Brysbaert, M.
\newblock (2013).
\newblock Norms of valence, arousal, and dominance for 13,915 english lemmas.
\newblock {\em Behavior research methods}, 45(4):1191--1207.

\end{thebibliography}

\section{Language Resource References}
\label{lr:ref}
\bibliographystylelanguageresource{lrec2022-bib}
\bibliographylanguageresource{languageresource}

\end{document}